\PassOptionsToPackage{table,dvipsnames}{xcolor}
\documentclass[sigconf,nonacm]{acmart}

\usepackage{enumitem}
\usepackage{multirow}
\usepackage{multicol}

\usepackage{booktabs}
\usepackage{array} 
\usepackage{caption} 

\usepackage{pifont}

\usepackage[most]{tcolorbox}
\usepackage[utf8]{inputenc}
\usepackage{tcolorbox}
\usepackage{tabularx}

\usepackage{ragged2e}
\usepackage[normalem]{ulem}
\usepackage{etoolbox}

\newcommand{\mwIItitleFullNameFirst}{Call of Duty\textregistered: Modern Warfare\textregistered II}
\newcommand{\mwIItitleFullName}{Call of Duty: Modern Warfare II}
\newcommand{\mwIItitleShort}{COD:MWII}

\newcommand{\wztitleFullNameFirst}{Call of Duty\textregistered: Warzone\texttrademark 2.0}
\newcommand{\wztitleFullName}{Call of Duty: Warzone 2.0}
\newcommand{\wztitleShort}{COD:WZ2}

\newcommand{\companyFirst}{Activision\textregistered{} Publishing, Inc.}
\newcommand{\company}{Activision}

\newcommand{\xmark}{\ding{55}} 

\newcommand{\seedDataset}{\textit{Initial Dataset}}
\newcommand{\deploymentDataset}{\textit{Deployment Dataset}}

\AtBeginDocument{%
  }

\begin{document}
\settopmatter{printacmref=false} 
\renewcommand\footnotetextcopyrightpermission[1]{} 

\title{Prosocial Behavior Detection in Player Game Chat: From Aligning Human-AI Definitions to Efficient Annotation at Scale}

\author{Rafal Kocielnik}
\email{rafalko@caltech.edu}
\orcid{0000-0001-5602-6056}
\affiliation{%
  \institution{California Institute of Technology}
  \city{Pasadena}
  \state{California}
  \country{USA}
}

\author{Min Kim}
\affiliation{%
  \institution{Activision Publishing, Inc.}
  \city{Santa Monica}
  \state{California}
  \country{USA}
}

\author{Penphob (Andrea) Boonyarungsrit}
\affiliation{%
 \institution{Activision Publishing, Inc.}
 \city{Santa Monica}
 \state{California}
 \country{USA}
}

\author{Fereshteh Soltani}
\affiliation{%
 \institution{Activision Publishing, Inc.}
 \city{Santa Monica}
 \state{California}
 \country{USA}
}

\author{Deshawn Sambrano}
\affiliation{%
 \institution{Activision Publishing, Inc.}
 \city{Santa Monica}
 \state{California}
 \country{USA}
}

\author{Animashree Anandkumar}
\affiliation{%
  \institution{California Institute of Technology}
  \city{Pasadena}
  \state{California}
  \country{USA}}

\author{R. Michael Alvarez}
\affiliation{%
  \institution{California Institute of Technology}
  \city{Pasadena}
  \state{California}
  \country{USA}}
\renewcommand{\shortauthors}{Kocielnik et al.}


\begin{abstract}
Detecting \textit{prosociality in text}—communication intended to affirm, support, or improve others’ behavior—is a novel and increasingly important challenge for trust and safety systems. Unlike toxic content detection, prosociality lacks well-established definitions and labeled data, requiring new approaches to both annotation and deployment. We present a practical, three-stage pipeline that enables scalable, high-precision prosocial content classification while minimizing human labeling effort and inference costs. First, we identify the best LLM-based labeling strategy using a small seed set of human-labeled examples. We then introduce a \textit{human-AI refinement loop}, where annotators review high-disagreement cases between GPT-4 and humans to iteratively \textit{clarify and expand the task definition}—a critical step for emerging annotation tasks like prosociality. This process results in improved label quality and definition alignment. Finally, we synthesize 10k high-quality labels using GPT-4 and train a \textit{two-stage inference system}: a lightweight classifier handles high-confidence predictions, while only $\sim$35\% of ambiguous instances are escalated to GPT-4o. This architecture reduces inference costs by \textasciitilde70\% while achieving high precision (\textasciitilde0.90). Our pipeline demonstrates how targeted human-AI interaction, careful task formulation, and deployment-aware architecture design can unlock scalable solutions for novel responsible AI tasks.
\end{abstract}       



\begin{teaserfigure}
  \includegraphics[width=\textwidth]{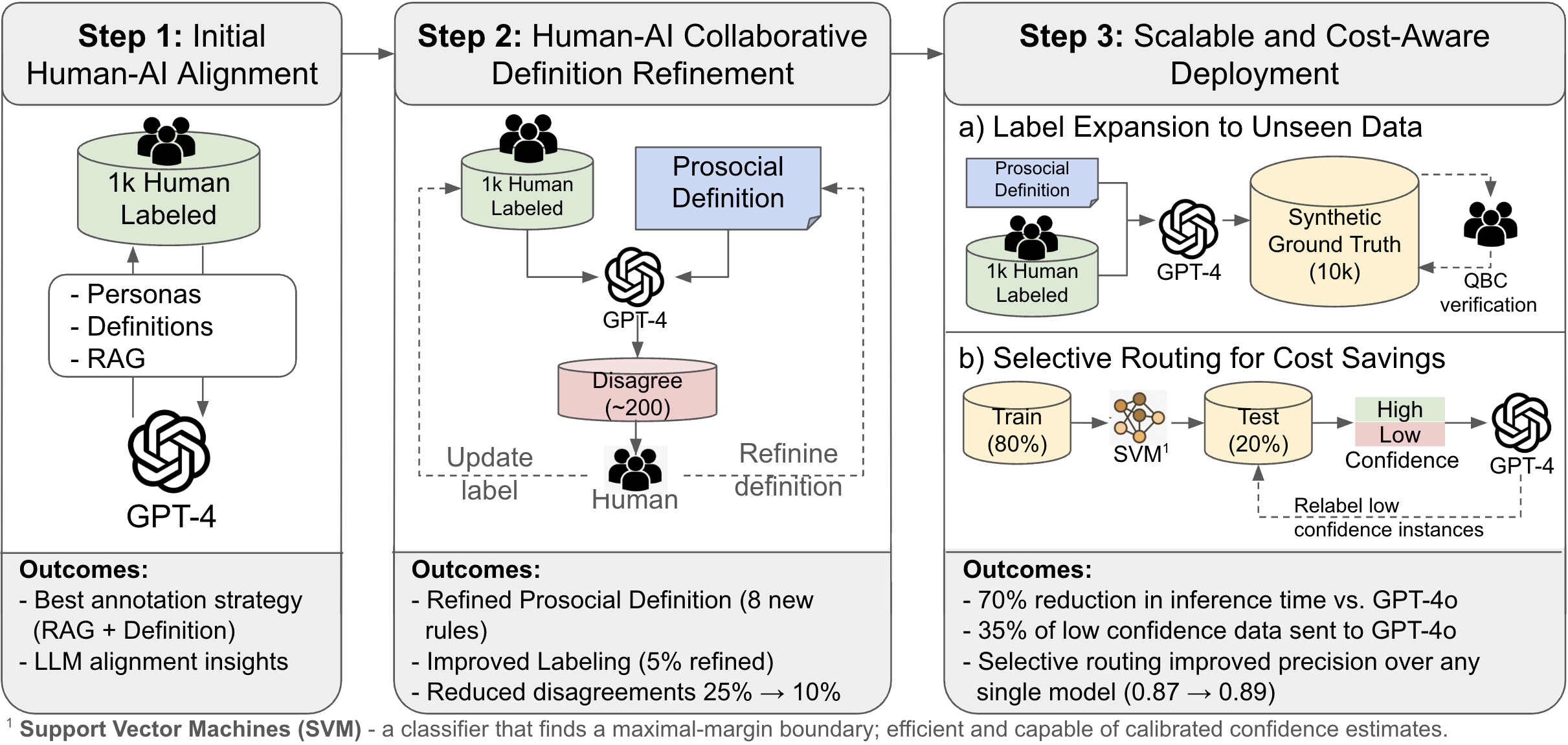}
  \caption{End-to-end pipeline for scalable, cost-efficient prosocial behavior classification in multiplayer game chat. 
\textbf{Step 1:} Evaluate annotation strategies on 1k human-labeled samples to align GPT-4 outputs. The best setup (RAG + definition) achieves AUC 0.85, Precision 0.93. 
\textbf{Step 2:} Refine definitions via $\sim$200 human-adjudicated disagreements, yielding 8 new rules, +5\% labeling improvement, and 25\%→10\% disagreement reduction. 
\textbf{Step 3a:} Label 10k unseen samples using refined GPT-4 prompts. A subset is audited via \textit{Query-by-Committee (QBC)} sampling to catch model disagreement. 
\textbf{Step 3b:} Train an SVM on synthetic labels. Only 35\% of uncertain predictions are routed to GPT-4, reducing inference cost by 70\% and improving precision (0.87→0.89).}
\label{fig:framework_overview}
\end{teaserfigure}


\maketitle

\section{Introduction}

\paragraph{Motivation.}
Positive social interactions are foundational to healthy gaming communities. While substantial effort has gone into detecting toxic behavior in online games, far less attention has been paid to identifying and amplifying \emph{prosocial} communication—interactions that promote cooperation, encouragement, or helpfulness. Yet, promoting prosociality is critical: recent work shows that rewarding positive behavior may be more effective at shaping player norms than punitive measures alone \cite{tomkinson2022thank}. This opens new opportunities for designing reward systems \cite{wijkstra2023help}, adaptive nudges \cite{nagle2014use}, and community-driven moderation tools that encourage supportive in-game communication \cite{beres2021don}.

\paragraph{Challenge.}
However, detecting prosocial behavior in game chat remains an underexplored and operationally difficult task. Unlike toxicity detection, which benefits from years of research \cite{wijkstra2023help, kordyaka2020towards}, shared benchmarks \cite{van2023unveiling, weld2021conda}, and pretrained models \cite{yang2023toxbuster, yang2025unified}, prosociality lacks (1) consistent, domain-relevant definitions in gaming, (2) labeled datasets for training reliable models, and (3) scalable annotation workflows suitable for production environments.

\paragraph{Our Contribution.}
We present a full-stack, deployment-aware pipeline for scalable prosocial behavior detection in real-world game chat (Figure \ref{fig:framework_overview}). This system was developed and evaluated in collaboration with domain experts at \companyFirst{} using data from a highly popular live competitive online action game titles - \emph{\mwIItitleFullNameFirst} and \emph{\wztitleFullNameFirst}. Our approach combines \emph{human-in-the-loop label refinement}, \emph{LLM-assisted annotation}, and a \emph{cost-efficient, hybrid classification system}. Concretely our framework involves 3 steps:
\begin{itemize}[leftmargin=*, itemsep=1pt, topsep=3pt]
    \item \textbf{Step 1: Initial Human-AI Alignment} We assess multiple LLM prompting strategies under a noisy seed definition of prosociality to align LLM-generated labels with human limited initial annotations. Retrieval-Augmented Generation (RAG) with definitions performs best in weakly supervised prosocial labeling.
    
    \item \textbf{Step 2: Human-AI Definition Refinement.} We introduce an adjudication loop between GPT-4o and expert human annotators to identify ambiguity, refine task definitions, and relabel a seed dataset using a collaborative process. This process helps align task framing with the domain context of game chat.

    \item \textbf{Step 3: Scalable and Cost-Aware Deployment.} We synthesize 10k additional labels using the refined prompt and train a lightweight Support Vector Machines (SVM) classifier with calibrated uncertainty \citep{wang2014survey}. At inference time, we use a \emph{two-stage system}: confident cases are classified by the SVM, while ambiguous ones ($\sim$35\%) are escalated to GPT-4o. This reduces LLM usage by $\sim$70\% while preserving high precision under  adjustable precision-recall-cost trade-off.
\end{itemize}

\paragraph{Deployment Insights.}
Our deployed solution provides four key takeaways for real-world LLM-assisted labeling systems:
\begin{itemize}[leftmargin=*, itemsep=1pt, topsep=3pt]
    \item \textbf{LLM outputs require domain grounding.} Without context-specific prompting, GPT-4o consistently mislabels competitive banter as toxic or non-prosocial—even falsely flagging phrases like \textit{``nice kill.''} Prompt vocabulary also matters: substituting label terms like \texttt{PROSOCIAL} for neutral placeholders skewed model predictions and precision.
    
    \item \textbf{Definition drift is a real risk.} Iteratively updating task prompts or definitions can introduce subtle semantic shifts that erode label consistency over time—underscoring the need for version control and structured review in human–LLM annotation loops.
    
    \item \textbf{Selective routing beats single-model baselines.} A calibrated SVM that escalates only $\sim$35 \% of uncertain cases to GPT-4o both cuts inference cost by close to 70 \% and yields higher precision than either model alone, illustrating that cost and quality can improve together.
\end{itemize}

\paragraph{Impact.}
This work represents the first large-scale deployment-oriented effort to detect prosocial behavior in game chat. It contributes both a generalizable LLM-human annotation pipeline and practical insights into aligning LLMs with underdefined social behaviors in high-volume, real-world environments. We believe this process can serve as a blueprint for responsibly scaling AI moderation tools that go beyond toxicity detection and actively promote positive interactions.

\section{Background and Related Work}

\subsection{Prosocial Behavior in Game Chat}

Emerging research has begun to explore prosocial communication in gaming environments, highlighting its potential for improving player experience and community health. For example, \citet{tomkinson2022thank} analyze expressions of gratitude across online gaming interactions, showing that such prosocial acts correlate with improved perceptions of teammate performance. Similarly, \citet{wijkstra2023help} conduct a large-scale analysis of in-game help-seeking and help-giving behavior, showing that such exchanges—though less frequent than toxic behavior—are widespread and socially impactful. These studies provide foundational evidence that prosocial behavior is not only present in gaming contexts, but meaningful for player outcomes.

However, existing work in this area is largely descriptive or retrospective in nature. Prior studies focus on identifying examples of prosociality post hoc and rely on manual annotation or rule-based heuristics rather than providing scalable, generalizable frameworks for detecting such behaviors in real-world systems. Furthermore, most prior research does not operationalize prosociality as a well-defined annotation task, nor does it address the lack of labeled data or the definitional ambiguity that often accompanies novel social constructs in chat data.

Our work builds on these insights by introducing a formal prosocial classification pipeline for in-game chat. We combine unsupervised pattern discovery, expert-in-the-loop definition refinement, and sample-efficient classification methods that scale to real-world datasets. Unlike prior studies, our approach explicitly confronts the challenges of definition development, label scarcity, and deployment feasibility—offering a full-stack methodology for detecting and ultimately incentivizing prosocial behavior in multiplayer game environments.

\subsection{LLM-Assisted Data Annotation}

Recent work has explored LLM-assisted strategies for labeling complex social data at scale. Farr et al.~\cite{farr2024mchain} propose a chain ensemble that routes data through multiple LLMs based on prediction confidence, combining outputs via rank-based aggregation to improve performance and reduce cost. However, their method assumes stable task definitions and does not incorporate human-in-the-loop refinement or concept disambiguation. Farr et al.~\cite{farr2024confidence} further assess uncertainty quantification methods to identify low-confidence predictions, but stop short of using these signals to iteratively refine task definitions or labels.

Other studies apply LLMs to specific social tasks. Horych et al.~\cite{horych2024promises} use GPT-4 to label media bias examples and train classifiers on these synthetic labels, assuming fixed annotation criteria and offering no feedback loop for improving label quality. Smith et al.~\cite{liu2025qualitative} focus on human-in-the-loop refinement of ambiguous annotations, improving agreement through guideline iteration, but do not scale beyond manual labeling or leverage LLMs.

Most closely related, Li et al.~\cite{li2025selfanchored} introduce a self-anchored attention model (SAAM) to detect prosocial chat using a small set of expert-labeled anchors. While their model improves low-resource classification performance, it does not leverage LLMs for scalable label synthesis, omit human-LLM refinement for definition development, and lacks a deployment-oriented inference strategy.

\subsection{Hybrid Inference and Cost-Aware Routing}
Recent studies have also begun to explore hybrid inference strategies that route low-confidence predictions to larger LLMs to balance accuracy and efficiency. Farr et al.~\cite{farr2024mchain} and others have demonstrated performance gains via confidence-based routing, but most approaches lack fine-grained control over recall-precision trade-offs or require multi-model orchestration. In contrast, our work uses a single lightweight model (SVM) with calibrated confidence to drive selective fallback to GPT-4o, enabling tunable routing without ensemble overhead.

In summary, while prior work on LLM-assisted annotation and confidence-aware inference provides useful building blocks, it often assumes well-defined tasks, static guidelines, or offline-only deployment contexts. Our work addresses these gaps by integrating iterative definition refinement, scalable synthetic labeling, and real-time hybrid inference into a unified pipeline tailored for social signal detection in online game chat or online chat more broadly.

\section{System Design}
We describe our methodology, including datasets used, human domain expert involvement, and finally our pipeline for LLM alignment, annotation refinement, and production scaling.

\subsection{Datasets}
We leveraged two different datasets, a small \seedDataset{} collected recently with noisy labels under an initial definition of prosociality adapted from \cite{li2025selfanchored}. The annotation of prosociality in this data was informed by general literature on prosocial behavior. We then collect and annotate \deploymentDataset{}, which we use for scaling-up and evaluation of our pipeline. Full descriptive statistics of these datasets are in Table \ref{tab:text_stats_datasets} in Appendix \ref{apx:dataset_stats}. Both datasets were de-identified.

\paragraph{\seedDataset{}:} We used a dataset collected in \citet{li2025selfanchored}, that was derived from a one-week sample of player text chat in the public channel of \mwIItitleFullName{} (\mwIItitleShort{}), collected between 2023-12-11 and 2023-12-17. It reflects real-time communication used by players during gameplay and was preprocessed to remove spam (e.g., emails, URLs), automatically flagged toxic content, and extremely short or low-variance entries to improve data quality. Each player chat instance represents a concatenation of all individual chat messages for a player during a match. They have an average length of 30.6 words (SD = 6.4), with a median of 32 words. The dataset was labeled by two independent human annotators following a prosociality definition based on prior literature. Disagreements between human raters were manually resolved via adjudication. The final labeled dataset includes 960 entries (53.0\% prosocial, 28.5\% not-prosocial, 18.4\% unclear), forming the basis for our iterative LLM alignment and definition refinement process. Further preprocessing details are available in \cite{li2025selfanchored},

\paragraph{\deploymentDataset{}:}
To evaluate the final refined prosociality definition and annotation process at scale, we collected a larger dataset from \textit{Battle Royale} and \textit{Demilitarized Zone} modes of \wztitleFullName{} (\wztitleShort{}) between 2024-09-02 and 2024-09-08. This yielded over 3 million player-match chat instances, which still represents just a subset of all the interactions. Similarly, all chat messages for a single player in a match were concatenated into one string with `.' as separator, unless a punctuation was already present. This was to preserve broader context and match the format of \seedDataset{} from prior work. The player chat instances in this dataset have an average length of 27.7 words (SD = 33.2), with a median of 18 words.
To focus on meaningful interactions, we filtered entries based on message length and spam characteristics, using word count buckets and thresholds for per-match message volume and duplication following procedures in \citep{li2025selfanchored}. We then applied the best-performing LLM configuration from Step 2 of our process in Figure \ref{fig:framework_overview} (RAG with refined definition) to label 10k new player-match chat entries. Manual inspection revealed nuanced labeling patterns: for example, 5.4\% of ``not-prosocial'' cases included the word ``good'' used sarcastically or competitively, while 12.4\% of ``prosocial'' instances used ``solo'' in contexts of seeking help or communication. Ultimately, only 27\% of the entries were labeled as prosocial, while 23\% were flagged as unclear.

\subsection{Domain Expert Involvement}
Our iterative definition refinement process (Step 2, Figure \ref{fig:framework_overview}) and system evaluation involved close collaboration with five gaming domain experts at \company{}. These individuals brought diverse perspectives spanning machine learning, behavioral analytics, gameplay experience, and moderation operations. Among them were a senior data analyst with deep expertise in player behavior and in-game communication, an NLP-focused machine learning lead with intimate familiarity with competitive game dynamics, and two avid \mwIItitleShort{} and \wztitleShort{} players involved in moderation and intervention design. Additional perspectives came from a mid-level product manager with experience running incentive and moderation experiments, and a moderation operations expert with hands-on experience in both automated and human-in-the-loop moderation systems.

These experts played a critical role in the human-AI collaboration process by reviewing high-disagreement cases between human annotations and GPT-4o predictions. They helped identify definitional ambiguities, surface domain-specific edge cases, and validate annotation strategies from both community norms and operational feasibility standpoints. This input ensured that the evolving prosocial definition remained behaviorally grounded, contextually relevant, and deployment-ready.

\subsection{Hardware \& Model Evaluation Setup}
All experiments were conducted on Azure cloud infrastructure. Classical and deep learning models were trained and evaluated on \texttt{a2-highgpu-1g} virtual machines with a single NVIDIA A100 GPU, 85~GB of memory, and 12 CPU cores. GPT-3.5, GPT-4-turbo, and GPT-4o were accessed in an inference-only setting via internal Azure OpenAI endpoints. Because these models run on private infrastructure, no data was transmitted externally or incorporated into commercial training sets. To avoid cross-instance contamination and data leakage~\citep{schroeder2024can}, all prompts were issued as separate API calls without batching, following best practices for judgment elicitation with LLMs~\citep{zheng2023judging}.

We partitioned the \deploymentDataset{} into an 80/20 split for development and held-out test sets. The development set was further divided into training and validation subsets using an 80/20 ratio. All splits were stratified to preserve the original label distribution. These partitions were used to train, validate, and select hyperparameters for trainable models including logistic regression, random forest, and shallow neural networks.

To ensure robust evaluation, all trainable models were run with three distinct random seeds. We report the mean performance across these runs. LLMs were evaluated using the same test set as other models to enable direct comparison, using zero-shot or few-shot prompting without any parameter updates.

\subsection{Annotation Refinement \& Scaling Process}
To enable scalable and cost-efficient prosocial labeling, we developed a structured annotation refinement and expansion pipeline combining human-in-the-loop review, GPT-4o-based label synthesis, and lightweight model deployment. This process began by aligning GPT-4o annotation behavior with human-labeled examples under an initial definition, then refined that definition iteratively through expert adjudication, and culminated in hybrid model deployment for efficient large-scale inference. We describe these in detail.

\begin{figure*}[t!]
  \centering
  \includegraphics[width=\textwidth]{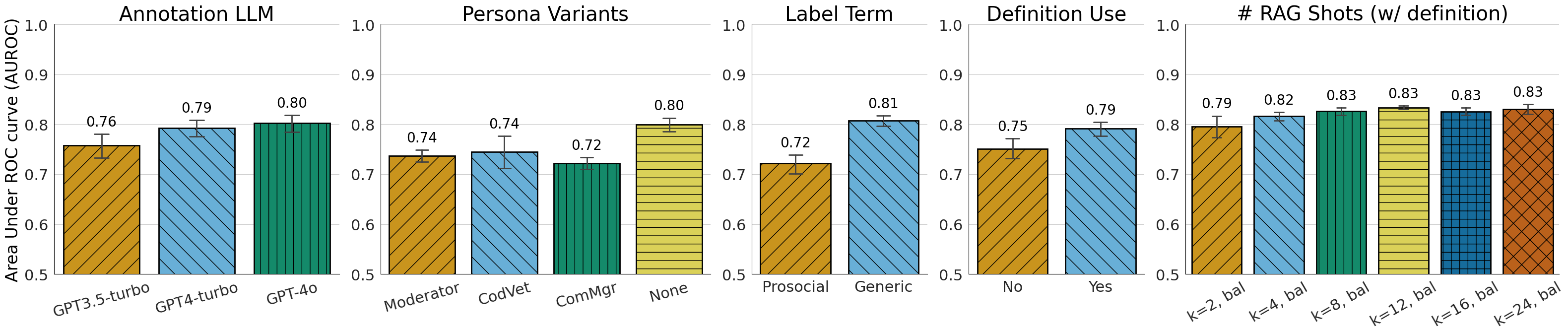}
  \vspace{-20pt}
  \caption{Impact of individual prompt design factors on LLM annotation performance on \emph{\seedDataset{}}, measured by Area Under the ROC Curve (AUROC). From left to right, the subplots compare: (1) different GPT model versions, (2) persona framings embedded in the prompt, (3) label terminology used (prosocial vs. generic), (4) inclusion vs. omission of a task definition, and (5) number of in-context RAG examples per class (balanced). Each bar shows the mean AUROC with 95\% confidence intervals, aggregated over five evaluation folds. Results indicate that prompt structure significantly affects annotation quality, with improvements linked to model choice, neutral framing, inclusion of task definitions, and the use of sufficient few-shot examples. These findings guided selection of the final labeling strategy used in our annotation pipeline.
}
  \label{fig:best_prompting_explore}
\end{figure*}

\paragraph{\textbf{Step 1: LLM Labeling Strategy Selection}}

As shown in Figure~\ref{fig:framework_overview}: Step 1, we initiated our annotation pipeline by aligning GPT-4o outputs with a seed dataset of $\sim$1k human-labeled chat messages, using an initial prosociality definition (D1) from \citet{li2025selfanchored} (see Appendix~\ref{apx:prosoc_def_evolution}, Table~\ref{tab:prosocial_definitions_a}). We systematically evaluated prompting strategies across three model versions (GPT-3.5-turbo, GPT-4-turbo, GPT-4o) and four core dimensions: (1) inclusion of task definitions (\textit{Yes/No}), (2) persona framing (\textit{Moderator, COD Veteran, Community Manager, or None}), (3) label terminology (\texttt{PROSOCIAL}/\texttt{NOT-PROSOCIAL} vs. \texttt{LABEL-1}/\texttt{LABEL-2}), and (4) the number of few-shot examples retrieved via retrieval-augmented generation (RAG) with $k \in \{2,\allowbreak 4,\allowbreak 8,\allowbreak 12,\allowbreak 16,\allowbreak 24\}$ per class, following \cite{kocielnik2023can, prabhumoye2021few}.

To support RAG, we designated an anchor set of 200 human-labeled examples from the seed dataset and computed cosine similarity between the embedded test instance and anchor candidates to retrieve the most relevant examples. Sentence embeddings were derived using models from the Sentence-Transformers library \citep{reimers-2019-sentence-bert}, with additional fine-tuning via supervised contrastive learning \citep{xu2023contrastive}. Among several embedding variants evaluated—\textit{``all-MiniLM-L6-v2''}, \textit{``all-MiniLM-L16-v2''}, \textit{``paraphrase-multilingual-MiniLM-L12-v2''}, and \textit{``GIST-small-Embedding-v0''}—the latter yielded the best retrieval performance and was used for all RAG experiments.

In total, our grid search spanned 66 unique prompting configurations: $3$ (models) $\times$ $2$ (definition inclusion) $\times$ $4$ (persona) $\times$ $2$ (label terms) + $3$ (models) $\times$ $6$ (RAG $k$-values), where RAG configurations were only evaluated in conjunction with definitions.

\emph{Persona framing} was motivated by prior work and grounded in gaming-domain roles. The \textit{Moderator} reflected social media moderation contexts \citep{sharma2025ai, kumar2024watch}; \textit{COD Veteran} was inspired by the League of Legends Tribunal system where experienced players adjudicate peer behavior \citep{blackburn2014stfu}; and \textit{Community Manager} mirrored professional roles managing player communications on social platforms \citep{garrelts2017responding}.

\emph{Label terminology} was included as a control for lexical priming: prior research suggests that LLMs may anchor behavior to semantically rich label tokens (e.g., \texttt{PROSOCIAL}), independent of contextual grounding \citep{fei2023mitigating}. To control for such effects, we compared prompts using domain-relevant versus generic class labels (\texttt{LABEL-1}/\texttt{LABEL-2}).

\emph{Few-shot example selection} followed a dynamic RAG approach rather than static inclusion. Prior work indicates that in-context examples retrieved via embedding similarity provide stronger steering signals for LLMs \citep{liu2024lost}. Our experiments varied the number of such examples to determine the optimal demonstration size for improving alignment with human-labeled data.

\begin{table}[hb!]
\centering
\caption{Comparison of selected top prompting configurations across models on \emph{\seedDataset{}}. Best AUC and precision values per group are highlighted. RAG (Retrieval-Augmented Generation) 4×2 and 16×2 refer to 4 or 16 examples per class (2 classes total) retrieved by cosine similarity as in-context examples.}
\vspace{-8pt}
\label{tab:prompting_comparison}
\begin{tabular}{llllllcc}
\toprule
\textbf{Persona} & \textbf{Def.} & \textbf{RAG} & \textbf{Label} & \textbf{GPT} & \textbf{AUC} & \textbf{Prec.} \\
\midrule
\multirow{3}{*}{\xmark} & \multirow{3}{*}{\checkmark} & \multirow{3}{*}{16$\times$2} & \multirow{3}{*}{Generic} 
    & 4o       & \textbf{0.85} & \textbf{0.93} \\
& & & & 4-turbo  & 0.82 & \textbf{0.93} \\
& & & & 3.5      & 0.81 & 0.79 \\
\midrule
\multirow{3}{*}{\xmark} & \multirow{3}{*}{\checkmark} & \multirow{3}{*}{4$\times$2} & \multirow{3}{*}{Generic} 
    & 4o       & 0.83 & 0.89 \\
& & & & 4-turbo  & 0.83 & 0.90 \\
& & & & 3.5      & 0.80 & 0.78 \\

\bottomrule
\end{tabular}
\end{table}

\paragraph{\textbf{Step 2: Human-AI Collaboration for Prosocial Definition Refinement}}

To improve alignment and reduce ambiguity in prosocial behavior detection, we implemented an iterative definition refinement process based on structured human-AI collaboration (Figure~\ref{fig:framework_overview}, Step~2). Human experts reviewed GPT-4o predictions on the \seedDataset{} to identify disagreements with the initial ground truth. Disagreements were manually analyzed to determine whether they resulted from annotation noise or unclear definition boundaries—both common challenges in subjective social behavior classification~\citep{chen2018using}. 

We adopted an adjudication protocol informed by qualitative coding practices~\citep{richards2018practical}, focusing on systematically reviewing model–human disagreements. Each disagreement instance was either used to refine the label or update the prosociality definition. Definition changes were first proposed by GPT-4o using a self-refinement prompting method~\citep{madaan2023self}, then reviewed and approved by expert raters. Label changes could result from either original mislabeling or re-interpretation under the updated definition.

The process continued iteratively until we achieved a satisfactory disagreement rate of under 10\% while maintaining precision above 0.95. This reflected deployment priorities: minimizing false positives (i.e., rewarding non-prosocial behavior) was deemed more costly than missing genuine prosocial examples. Thus, the definition was optimized for high-precision classification under ambiguity.

In addition to aligning labels, the goal was to produce a behaviorally grounded, operationally clear definition of prosociality in competitive gaming text chat—supporting both model generalization and human interpretability at scale. This approach draws from collaborative coding systems in HCI and computational social science~\citep{drouhard2017aeonium, vila2024abductive}.

\begin{table*}[t!]
\centering
\caption{Summary of definition refinement across six iterations on \emph{\seedDataset{}}. Each row shows how changes to the prosociality definition affected LLM-human disagreement, label stability, and semantic content. Disagreement consistently decreased as definitions became more specific, structured, and aligned with annotation needs. Exact definition with highlighted changes are in Tables \ref{tab:prosocial_definitions_a} and \ref{tab:prosocial_definitions_b} in Appendix \ref{apx:prosoc_def_evolution}.}
\vspace{-10pt}
\label{tab:def_refinement}
\begin{tabular}{c c c c c p{8.0cm}} 
\toprule
\textbf{Def. ID} & \textbf{\# Disagr.} & \textbf{\% Label Chg.} & \textbf{\% Def. Chg.} & \textbf{Final Disagr. \%} & \textbf{Notable Edit} \\
\midrule
D1 & 218 & 0\% & 0\% & 25.4\% & \textbf{Initial definition} based on general-purpose framing of prosociality in broader contexts. \\
D2 & 177 & 0\% & 91\% & 22.4\% & \textbf{Reframed} from intent-based to outcome-based; added \textbf{structured categories} and in-game examples. \\
D3 & 163 & 2\% & 5\% & 19.3\% & \textbf{Introduced exclusions} for vague help-seeking and team critique to reduce ambiguity. \\
D4 & 124 & 2\% & 8\% & 14.7\% & \textbf{Added anti-patterns} (e.g., toxicity, threats); \textbf{reinstated equipment-help} cases. \\
D5 & 118 & 1\% & 3\% & 12.8\% & \textbf{Refined language} for action clarity and cleaned up exclusion phrasing. \\
D6 & 93 & 1\% & 9\% & 9.7\% & \textbf{Finalized exclusions} (e.g., nagging, complaints); clarified \textbf{support-related cases}. \\
\bottomrule
\end{tabular}
\end{table*}

\paragraph{\textbf{Step 3: Label Expansion \& Cost Efficient Labeling at Scale}}

To scale classification, we applied the optimal GPT-4o prompting configuration—specifically, a 16$\times$2 retrieval-augmented generation (RAG) setup incorporating the refined prosociality definition without persona priming—to annotate 10{,}000 previously unseen chat instances from \deploymentDataset{} (Figure~\ref{fig:framework_overview}, Step~3). These high-fidelity synthetic labels served as supervision for training a suite of classical and neural models evaluated across performance, calibration, and inference efficiency dimensions.

To ensure annotation quality, we adopted a human-in-the-loop refinement process inspired by prior work on model disagreement resolution~\citep{tan2024large, horych2024promises}. Annotators manually reviewed disagreements between GPT-4o and other model predictions using the updated prosociality definition, yielding an adjudicated ground truth set used in all evaluations. GPT-4o was also reapplied with the same prompting strategy to benchmark its own performance on this new gold standard.

We trained and benchmarked a variety of models using the synthetic labels. Classical classifiers included logistic regression, random forest, and SVM, while neural models included LSTM, SetFit, and BERT. Each model was assessed in terms of predictive performance, calibration (via Expected Calibration Error, ECE \citep{posocco2021estimating}), and inference cost. GPT-4o was reserved for labeling and high-precision fallback, not direct model training, due to cost and latency concerns.

To reduce dependency on costly LLM inference, we implemented a two-stage hybrid routing system. The base classifier—a calibrated SVM—was used to confidently classify high-certainty examples. Predictions falling within a tunable uncertainty window, defined by a midpoint (e.g., 0.3, 0.5, or 0.7) and a tolerance interval around that midpoint (i.e., midpoint ± tolerance), were routed to GPT-4o. This architecture supports configurable trade-offs between performance and computational cost by adjusting the routing threshold and fallback frequency dynamically.

\section{Evaluation and Results}

\subsection{Step 1: LLM Labeling Strategy Selection}

\begin{figure}[ht]
  \centering
  \small
  \includegraphics[width=\columnwidth]{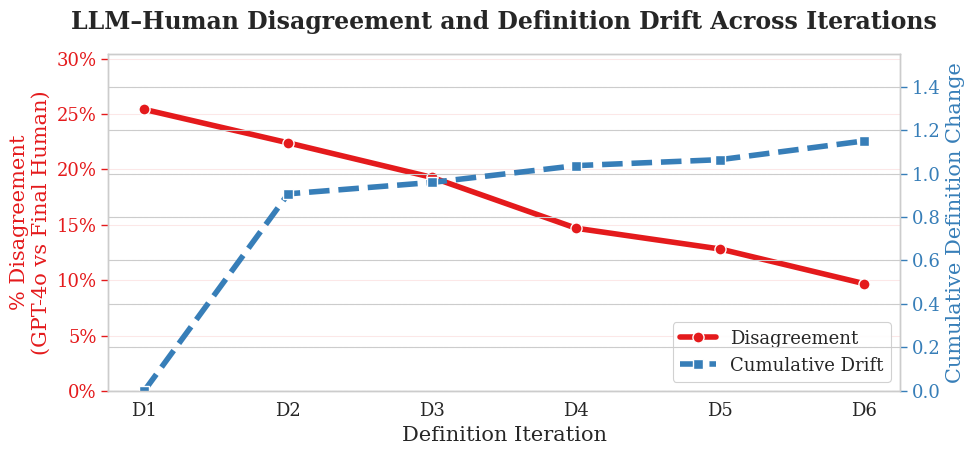} 
  \vspace{-20pt}
  \caption{
  \textbf{LLM–human disagreement and definition drift (D1–D6).}
  \textcolor{red}{\textbf{Red}} = disagreement rate; \textcolor{blue}{\textbf{blue}} = cumulative drift (1 – cos sim). Disagreement falls from 25 \% to 10 \% as successive edits -- outcome framing (D2), exclusions (D3), and anti-patterns (D4) -- align the model with human labels.
}
  \label{fig:disagreement_drift}
  \vspace{-6pt}
\end{figure}

The best-performing setup combined Retrieval-Augmented Generation (RAG) with 16 examples per class, the inclusion of a task definition, use of a generic label term, and no persona framing, achieving an AUROC of 0.85 and a precision of 0.93 with GPT-4o (Table~\ref{tab:prompting_comparison}). To understand the contribution of individual components, we conducted a series of controlled comparisons across model variants, persona framing, label terminology, definition use, and the number of in-context examples (Figure~\ref{fig:best_prompting_explore}). We observed that newer model versions consistently outperformed earlier ones, with GPT-4o yielding slightly higher AUROC than GPT-4-turbo and GPT-3.5-turbo (0.80 vs. 0.79 and 0.76, respectively). Removing persona framing led to a substantial performance increase (AUROC: 0.80), while persona variants such as \textit{``Moderator''} or \textit{``Community Manager''} reduced performance, suggesting that role-based priming may introduce unintended biases. Similarly, using a generic label term outperformed task-specific phrasing such as \texttt{PROSOCIAL} (AUROC: 0.81 vs. 0.72), likely due to reduced semantic anchoring or priming effects. Including a concise task definition improved performance (AUROC: 0.79 vs. 0.75 without definition), supporting prior findings that clear task framing aids LLM alignment. Finally, increasing the number of in-context RAG examples improved annotation quality up to 8–16 per class (AUROC plateauing at 0.83), with no further gains beyond that. These results highlight that small adjustments in prompt structure—particularly in terms of framing neutrality, clarity, and example selection—can significantly affect the reliability and alignment of LLM annotations.

\begin{figure*}[t!]
  \centering
  \caption{
  \textbf{Trade-off between routing threshold tolerance, performance, and inference cost across uncertainty midpoints on \emph{\deploymentDataset{}}.} 
  Each line shows the effect of varying tolerance (x-axis) around an SVM probability midpoint (0.3, 0.5, 0.7) on (1) the percentage of data routed to GPT-4o (left), (2) ROC AUC, (3) precision, and (4) recall. As tolerance increases, more data is routed, improving overall performance but at higher inference cost. A midpoint near 0.3 reroutes many low-score (borderline) cases, boosting recall; one near 0.7 reroutes only high-score cases, boosting precision.
  }
  \includegraphics[width=\textwidth]{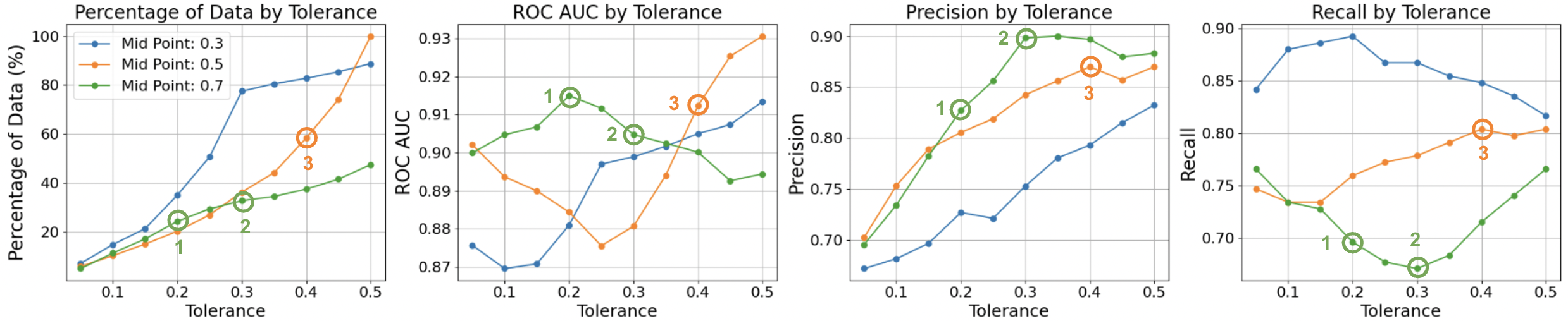}
  
  \label{fig:routing_tredeoffs}
\end{figure*}

\begin{table*}[t]
\centering
\vspace{-4pt}
\caption{Performance and inference cost at varying data routing percentages to GPT-4o on \emph{\deploymentDataset{}}, grouped by SVM probability midpoint selection (corresponds to Figure \ref{fig:routing_tredeoffs}). Routing larger proportions of uncertain data increases cost while generally improving precision and recall. The 0.3 midpoint around SVM probability scores optimizes Recall over Precision at the lower \% of data routed. Top 3 recall values are highlighted in \raisebox{0pt}[0pt][0pt]{\colorbox{blue!20}{\rule{0pt}{0.1ex}\hspace{0.0em}blue\hspace{0.0em}}}. On the other hand, the 0.7 midpoint optimizes Precision over Recall at the lower \% of data routed. Top 3 precision values are highlighted in \raisebox{0pt}[0pt][0pt]{\colorbox{ForestGreen!20}{\rule{0pt}{0.1ex}\hspace{0.0em}green\hspace{0.0em}}}.}

\vspace{-8pt}

\label{tab:routing_tradeoffs_fine}
\begin{tabular}{llccccccccccc}
\toprule
& & \multicolumn{11}{c}{\textbf{\% Data Routed from SVM to GPT-4o Based on Uncertainty Estimate}} \\
\cmidrule(lr){3-13}
\textbf{Midpoint} & \textbf{Metric} & \textbf{0\%} & \textbf{10\%} & \textbf{20\%} & \textbf{30\%} & \textbf{40\%} & \textbf{50\%} & \textbf{60\%} & \textbf{70\%} & \textbf{80\%} & \textbf{90\%} & \textbf{100\%} \\
\midrule
& Inference Cost (sec) & 0.0015 & 0.1933 & 0.3885 & 0.5837 & 0.7788 & 0.9740 & 1.1692 & 1.3644 & 1.5596 & 1.7548 & 1.9500 \\
\midrule

\multirow{2}{1.7cm}{0.3 (\textuparrow Recall)}
& Precision & 0.670 & 0.677 & 0.697 & 0.714 & 0.723 & 0.721 & 0.730 & 0.746 & 0.776 & 0.838 & 0.870 \\
& Recall & 0.785 & 0.861 & \cellcolor{blue!15}0.886 & \cellcolor{blue!15}0.886 & \cellcolor{blue!15}0.892 & 0.867 & 0.873 & 0.873 & 0.854 & 0.817 & 0.804 \\
\midrule

\multirow{2}{1.7cm}{0.7 (\textuparrow Precision)}
& Precision & 0.670 & 0.718 & 0.792 & 0.856 & \cellcolor{ForestGreen!20}0.892 & \cellcolor{ForestGreen!20}0.884 & \cellcolor{ForestGreen!20}0.887 & 0.875 & 0.869 & 0.870 & 0.870 \\
& Recall & 0.785 & 0.741 & 0.722 & 0.677 & 0.734 & 0.772 & 0.791 & 0.798 & 0.798 & 0.804 & 0.804 \\
\bottomrule
\vspace{-16pt}

\end{tabular}
\end{table*}

\subsection{Step 2: Human-AI Collaboration for Prosocial Definition Refinement}
We conducted six iterations of definition refinement, guided by model–human disagreement analysis. At each step, we tracked: (1) the number of GPT-4o vs. human disagreements, (2) the proportion of label changes (\% Label Chg.), (3) the degree of semantic definition change via 1 – cosine similarity over TF-IDF embeddings (\% Def. Chg.), and (4) final disagreement against gold-standard labels.

As shown in Figure~\ref{fig:disagreement_drift} and Table~\ref{tab:def_refinement}, disagreement declined most sharply at targeted update points. Iteration D2 introduced outcome-based framing and behavioral categories, reducing disagreement from 25.4\% to 22.4\%. D3 added exclusion criteria (dropping to 19.3\%), and D4 introduced explicit anti-patterns (further reducing to 14.7\%). These targeted refinements consistently improved label consistency without model retraining. 

Overall, disagreement was reduced by 63\%—from 218 instances (25.4\%) in D1 to 93 (9.7\%) in D6—demonstrating that structured, iterative updates guided by human-LLM disagreements can enhance annotation consistency and clarify semantic boundaries.

\subsection{Step 3: Label Expansion \& Cost Efficient Labeling at Scale}
\subsubsection{Benchmarking Prediction Performance}
We evaluated a range of classical and neural models on the 10k high-fidelity synthetic labels generated by GPT-4o and refined via human review. As shown in Table~\ref{tab:model_comparison}, classical models delivered strong performance at low computational cost. Logistic regression and random forest achieved AUCs of 0.76 with inference times of 0.45–2.13s per 2,000 instances. The SVM outperformed both (AUC = 0.77, precision = 0.67, recall = 0.78), with moderate inference latency (2.93s).

Neural models—LSTM, SetFit, and BERT—offered improved predictive performance (BERT AUC = 0.81, precision = 0.76, recall = 0.79), but incurred higher inference latency (up to 180s per 2k instances). GPT-4o, used in inference, achieved the highest AUC (0.90, precision = 0.87, recall = 0.80) with 16$\times$2 RAG prompting, though inference cost exceeded one hour for 2k examples. These results illustrate the trade-off between model complexity and inference efficiency, motivating the hybrid routing strategy detailed below.

\subsubsection{Assessing Uncertainty Calibration}
Reliable uncertainty estimates are essential for routing borderline cases in hybrid systems. We computed Expected Calibration Error (ECE) to assess alignment between predicted probabilities and actual correctness. As shown in Appendix~\ref{apx:calibration_vsm_bert} (Figure~\ref{fig:apx_calib_plots}), SVM exhibited strong calibration (ECE = 0.0352), while BERT showed overconfidence and misalignment in mid-probability regions (ECE = 0.0993), consistent with prior findings~\citep{fakour2024structured}. Based on these results, we adopted SVM as the routing base model.

\subsubsection{Hybrid Routing Model and Tolerance Tuning}

To reduce reliance on expensive GPT-4o inference, we deployed a hybrid routing system. The calibrated SVM handles high-confidence predictions directly, while low-confidence cases---falling within a configurable uncertainty window---are routed to GPT-4o. This uncertainty window is defined by a midpoint (e.g., 0.3, 0.5, or 0.7) and a tolerance value $\epsilon$, such that predictions with scores in the range $[\text{midpoint} - \epsilon,\ \text{midpoint} + \epsilon]$ are rerouted.

Figure~\ref{fig:routing_tredeoffs} illustrates how varying tolerance affects performance and cost across different midpoints. At midpoint 0.3, the routing window captures lower-confidence predictions, increasing recall (up to 0.89) as more data is routed. In contrast, midpoint 0.7 focuses on higher-confidence cases, enabling precision gains (up to 0.89) while maintaining moderate routing volume. We highlight three illustrative trade-off points: \textbf{(1)} maximizing AUROC ($\sim$0.92) under low routing volume (green, midpoint 0.7, tolerance 0.2, $\sim$25\% of data), \textbf{(2)} maximizing precision ($\sim$0.90\%) under moderate routing (green, midpoint 0.7, tolerance 0.3, $\sim$35\% of data), and \textbf{(3)} balancing both precision ($\sim$0.87\%) and recall ($\sim$0.80\%) under higher routing (orange, midpoint 0.5, tolerance 0.4, $\sim$60\% of data).

As summarized in Table~\ref{tab:routing_tradeoffs_fine}, routing decisions based on midpoint and tolerance directly shape model behavior. For example, routing 30--40\% of data using a midpoint of 0.3 boosts recall substantially (to 0.89) with modest precision gains. In contrast, midpoint 0.7 with 40--50\% routing maximizes precision (up to 0.89), with stable recall around 0.77--0.80. Notably, this exceeds the precision of any standalone model in Table~\ref{tab:model_comparison}, including GPT-4o itself (0.87). Inference cost grows linearly with routing, from near-zero (SVM-only) to 1.95s per instance at full GPT-4o deferral.

This finding underscores the effectiveness of selective rerouting: by leveraging GPT-4o only for uncertain predictions, the hybrid system not only reduces latency but also outperforms all individual models in terms of precision.

\begin{table}[t]
\centering
\caption{Comparison of model performance and efficiency for prosociality classification on \emph{\deploymentDataset{}}. Simpler models are faster to train and infer but deep learning models improve classification. GPT-4o provides high-fidelity and flexible labeling but at significant latency cost.}
\vspace{-10pt}
\label{tab:model_comparison}
\begin{tabular}{l c c c c c}
\toprule
\textbf{Model} & \textbf{AUC} & \textbf{Prec.} & \textbf{Rec.} & \textbf{Train time} & \textbf{Test time*} \\
\midrule
\multicolumn{6}{l}{\textit{Classical ML models}} \\
\midrule
Logistic & 0.76 & 0.65 & 0.82 & 4.30 sec & 0.45 sec \\
Rnd. Forest & 0.76 & 0.63 & 0.76 & 15.73 sec & 2.13 sec \\
SVM & 0.77 & 0.67 & 0.78 & 76.61 sec & 2.93 sec \\
\midrule
\multicolumn{6}{l}{\textit{Neural models}} \\
\midrule
LSTM & 0.78 & 0.70 & 0.79 & 76.34 sec & 2.26 sec \\
SetFit\textsuperscript{1} & 0.80 & 0.76 & 0.78 & $\sim$39 min & 1.21 sec \\
BERT & 0.81 & 0.76 & 0.79 & $\sim$36 min & $\sim$180.0 sec \\
\midrule
GPT-4o\textsuperscript & 0.90 & 0.87 & 0.80 & -- & $\sim$4000 sec \\
\bottomrule
\end{tabular}

\vspace{0.5em}
\raggedright
\footnotesize
\textit{*Inference time measured over 2000 player chat instances.} \\
\textsuperscript{1} SetFit: Efficient few-shot fine-tuning with sentence transformers. \\
\vspace{-10pt}
\end{table}

\vspace{-2pt}
\section{Discussion}

Our study shows that a \emph{selective-routing} architecture can achieve improved precision while cutting large-model usage by close to 70\%.  Conceptually, the calibrated SVM acts as a risk-aware abstainer in the sense of selective-classification frameworks \citep{el2010foundations}, forwarding only low-confidence items to GPT-4o. Compared with heavyweight mixture-of-experts systems \citep{du2022glam}, this single-fallback design imposes minimal orchestration overhead yet still surpasses the oracle’s stand-alone precision (Table~\ref{tab:routing_tradeoffs_fine}).  From a deployment standpoint, it offers an attractive middle ground between fully LLM reliant moderation and purely local models: latency remains under 1.13 seconds per 1k instances for $\sim$75 \% of traffic, while the long-tail cases receive GPT-4o’s richer reasoning.

Equally important is our \emph{definition-engineering loop}. Six guided revisions of the prosociality guideline reduced human–model disagreement from 25 \% to under 10 \% (Figure~\ref{fig:disagreement_drift}) with no retraining.  Treating the prosociality definition as an optimizable artifact echoes recent data-centric work around LLM prompt optimization \citep{khattab2024dspy} as well as more human-centric work in qualitative research around Human-AI collaboration \citep{schroeder2025large}, but also highlights the operational danger of \emph{definition drift}: automated prompt tweaks that seem innocuous can silently shift label semantics and erode precision. By version-controlling every textual change and expert human adjudication oversight step, we provide an auditable trail consistent with emerging recommendations for dynamic policy governance \citep{nookala2024adaptive}.

Finally, our prompt study uncovers a pronounced \emph{semantic-anchoring} effect: replacing the value-laden token \texttt{PROSOCIAL} with a neutral placeholder improves AUROC by up to 8 points (Figure~\ref{fig:best_prompting_explore}). This finding complements controlled experiments on label-name bias in instruction-tuned models \citep{fei2023mitigating} and resonates with the priming failures reported by \citet{han2024context}. Likewise, role prompts such as \textit{Moderator} lower performance—potentially because they invoke an enforcement-focused moderation schema that clashes with the desired altruistic stance \citep{van2024speaking}. Routine counter-prompt audits such as those proposed by \citet{wang2023large} should therefore accompany any large-scale LLM annotation effort.

\subsubsection{Future Work}
There are several promising directions for extending our pipeline. First, testing the system in different settings—like moving from a fast-paced FPS to a cooperative MMO or even live-stream chat—could help us understand how different types of language, slang, and style-shifting (code-switching) affect the model’s performance. Second, we could explore ways to make the definition refinement process more efficient by using tools that automatically suggest rule changes, which would reduce the workload for human experts while keeping the model's reliability. Third, swapping out GPT-4o for a smaller, optimized language model (like a quantized 7B–11B model) with fast decoding methods could make the system much faster and potentially usable for real-time chat, including voice. Lastly, if we integrated the classifier into a live in-game system that rewards positive behavior, we could study whether this actually leads to more prosocial behavior over time.

\section*{Ethical Considerations}
All data used in this study originates from de-identified in-game chat logs collected under existing moderation protocols, with no personally identifiable information (PII) retained. The initial seed dataset was annotated by domain experts, while large-scale labeling was performed using language models (LLMs) via prompting without incorporating any sensitive or private content. We ensured that no training data was sent to external services: all LLM inference was conducted using internal Azure OpenAI infrastructure, guaranteeing that user data was not transmitted outside organizational boundaries or incorporated into commercial model training pipelines. 

Regarding player consent, we note that players were not enrolled in any additional experiments beyond normal gameplay. The collection and use of communication data for moderation and research purposes falls under Activision’s existing Terms of Use, which includes a \textit{``consent to monitor''} clause~\footnote{\url{https://www.activision.com/legal/terms-of-use}}. 

To mitigate potential labeling bias, we employed a human-in-the-loop definition refinement process and conducted adjudication on model disagreements. We adhered to best practices in responsible annotation, prompt design, and evaluation, and followed recent guidelines on LLM-based judgment elicitation~\citep{zheng2023judging}. No automated outputs were used in production or exposed to end users. This study received internal approval and was conducted in alignment with institutional data governance policies.

\vspace{-2pt}
\section{Conclusions}
This work introduces a scalable, deployment-ready pipeline for detecting prosocial behavior in multiplayer game chat—a task traditionally underdefined and data-scarce. By combining definition engineering through human-AI collaboration, LLM-assisted labeling, and hybrid cost-aware inference, our approach achieves production-level precision while reducing reliance on large models by nearly 70\%. Central to our success is the iterative refinement of task definitions based on human–LLM disagreement, highlighting the need for data-centric iteration even in high-performing systems. Our selective-routing architecture—using a calibrated SVM as a risk-aware abstainer—outperforms standalone LLMs in precision, offering a practical blueprint for content moderation systems that go beyond toxicity detection. This work advances the applied data science of responsible AI deployments and opens new pathways for reinforcing prosocial norms through AI-assisted feedback, moderation, and community design.

\begin{acks}
We would like to thank Michael Vance, Gary Quan, and Sabrina Hameister for their assistance with this project.

\end{acks}

\bibliographystyle{ACM-Reference-Format}
\bibliography{main_arxiv}

\appendix

\section{Datasets Descriptive Statistics}
\label{apx:dataset_stats}

Table~\ref{tab:text_stats_datasets} summarizes descriptive statistics for the two datasets used in this study: the seed dataset (\seedDataset{}) and the deployment-scale dataset (\deploymentDataset{}). We report the distribution of message lengths in terms of both word and character counts, disaggregated by prosociality labels. These statistics provide insight into the linguistic variability and distributional characteristics across label classes, informing modeling and evaluation design choices discussed in the main paper.

\begin{table*}[t]
\centering
\caption{Text length statistics for \seedDataset{} and \deploymentDataset{} by prosociality label. Word and character length distributions are reported with minimum, maximum, mean, median, and standard deviation values.}
\label{tab:text_stats_datasets}
\begin{tabular}{@{}p{3.2cm}lcccc@{\hspace{1cm}}cccc@{}}
\toprule
\textbf{Statistic} & \multicolumn{4}{c}{\textbf{Initial Dataset}} & \multicolumn{4}{c}{\textbf{Deployment Dataset}} \\
\cmidrule(lr){2-5} \cmidrule(lr){6-9}
& \textbf{All} & \textbf{Prosocial} & \textbf{Unclear} & \textbf{Non-prosocial}
& \textbf{All} & \textbf{Prosocial} & \textbf{Unclear} & \textbf{Non-prosocial} \\
\midrule
\textbf{Count} & 960 & 509 & 177 & 274 & 10000 & 2667 & 2305 & 5028 \\
\textbf{Percent} & (100\%) & (53.0\%) & (18.4\%) & (28.5\%) & (100\%) & (26.7\%) & (23.0\%) & (50.3\%) \\
\addlinespace
\multicolumn{9}{@{}l}{\textbf{Word Length}} \\
Min & 11.0 & 11.0 & 11.0 & 12.0 & 1.0 & 1.0 & 5.0 & 5.0 \\
Max & 39.0 & 39.0 & 39.0 & 39.0 & 493.0 & 389.0 & 493.0 & 438.0 \\
Mean & 30.6 & 30.6 & 29.9 & 30.9 & 27.7 & 25.8 & 32.4 & 26.5 \\
Median & 32.0 & 32.0 & 31.0 & 32.0 & 18.0 & 18.0 & 17.0 & 18.0 \\
Std & 6.4 & 6.4 & 7.0 & 6.1 & 33.2 & 28.5 & 46.3 & 27.6 \\
\addlinespace
\multicolumn{9}{@{}l}{\textbf{Character Length}} \\
Min & 49.0 & 52.0 & 49.0 & 66.0 & 7.0 & 7.0 & 18.0 & 17.0 \\
Max & 746.0 & 336.0 & 589.0 & 746.0 & 3066.0 & 2341.0 & 3066.0 & 2287.0 \\
Mean & 160.4 & 157.9 & 158.6 & 166.1 & 137.9 & 126.3 & 162.1 & 132.9 \\
Median & 161.0 & 159.0 & 157.0 & 165.0 & 88.0 & 85.0 & 84.0 & 91.0 \\
Std & 47.0 & 37.5 & 56.1 & 55.5 & 170.4 & 143.8 & 240.5 & 141.1 \\
\bottomrule
\end{tabular}
\end{table*}

\section{Human-AI Definition Refinement Process}
\label{apx:prosoc_def_evolution}

To support high-quality annotation and model evaluation, we iteratively refined the operational definition of prosocial behavior across six versions (D1–D6). Each version incorporated structured edits informed by observed LLM-human disagreements, annotation ambiguity, and edge cases encountered during labeling. 

The full text of each version is presented in Table~\ref{tab:prosocial_definitions_a} and Table~\ref{tab:prosocial_definitions_b}, capturing the progression from general intent-based framing to behaviorally specific, exclusion-aware guidelines.

\begin{table*}[ht]
  \centering
  \caption{\textbf{Evolution of the Definition of Prosocial Behavior Used in Evaluation.} Cumulative added exclusions are marked in \textcolor{red}{red} and inclusions in \textcolor{ForestGreen}{green}. Changes made from precious definition iteration are \underline{undelined}.}
  \vspace{0.5em}
  \label{tab:prosocial_definitions_a}
  \renewcommand{\arraystretch}{1.4}
  \begin{tabularx}{\textwidth}{>{\bfseries}l|X}
    \toprule
    Def. ID & Description \\
    \midrule
    D1 & 
    \begin{tcolorbox}[
      colback=white,
      colframe=white,
      boxrule=0.3pt,
      arc=0pt,
      outer arc=0pt,
      left=4pt, right=4pt, top=4pt, bottom=-6pt,
      sharp corners
    ]
    An interaction or behavior that is intended to results in benefit to another player. This includes, but is not limited, to behaviors such as helping and defending, assisting or acting in someone else’s interest, volunteering, donating, and comforting. As well as cooperation and sharing, expressing empathy and emotional awareness.
    \end{tcolorbox}
    \\
    \midrule
    D2 & 
    \begin{tcolorbox}[
      colback=white,
      colframe=white,
      boxrule=0.3pt,
      arc=0pt,
      outer arc=0pt,
      left=4pt, right=4pt, top=4pt, bottom=-6pt,
      sharp corners
    ]
    Any action or interaction that benefits other players or the community by fostering a positive gaming environment.\\
    Non-exhaustive list of examples: \\
    \textbf{Cooperating with other players:} e.g., strategizing with team, spawning with team, assisting, pinging for valuable loot or nearby enemies, defending the team.\\
    \textbf{Sharing with other players:} e.g., resources, revives, rewards/gifts.\\
    \textbf{Helping other players:} e.g., providing guidance to a less experienced player, positive reinforcement (verbal/written encouragement or compliments).\\
    \textbf{Helping the community:} e.g., encouraging others to report disruptive behavior, submitting validated reports for disruptive behavior, standing up for another player.\\
    \textbf{Celebrating with other players}\\
    \textbf{Congratulating other players}\\
    \textbf{Thanking other players}
    \end{tcolorbox}
    \\
    \midrule
    D3 & 
    \begin{tcolorbox}[
      colback=white,
      colframe=white,
      boxrule=0.3pt,
      arc=0pt,
      outer arc=0pt,
      left=4pt, right=4pt, top=4pt, bottom=-6pt,
      sharp corners
    ]
    Any action or interaction that benefits other players or the community by fostering a positive gaming environment.\\
    Non-exhaustive list of examples: \\
    \textbf{Cooperating with other players:} e.g., strategizing with team, spawning with team, assisting, pinging for valuable loot or nearby enemies, defending the team.\\
    \textbf{Sharing with other players:} e.g., resources, revives, rewards/gifts.\\
    \textbf{Helping other players:} e.g., providing guidance to a less experienced player, positive reinforcement (verbal/written encouragement or compliments).\\
    \textbf{Helping the community:} e.g., encouraging others to report disruptive behavior, submitting validated reports for disruptive behavior, standing up for another player.\\
    \textbf{Celebrating with other players}\\
    \textbf{Congratulating other players}\\
    \textbf{Thanking other players}\\
    \textcolor{red}{
    \underline{Exclude only requesting help and support without offering any.} \\
    \underline{Exclude simply pointing out actions that can negatively affect team cohesion.}}
    
    \end{tcolorbox}
    \\
    \midrule
    D4 & 
    \begin{tcolorbox}[
      colback=white,
      colframe=white,
      boxrule=0.3pt,
      arc=0pt,
      outer arc=0pt,
      left=4pt, right=4pt, top=4pt, bottom=-6pt,
      sharp corners
    ]
    Any action or interaction that benefits other players or the community by fostering a positive gaming environment.\\
    Non-exhaustive list of examples: \\
    \textbf{Cooperating with other players:} e.g., strategizing with team, spawning with team, assisting, pinging for valuable loot or nearby enemies, defending the team.\\
    \textbf{Sharing with other players:} e.g., resources, revives, rewards/gifts.\\
    \textbf{Helping other players:} e.g., providing guidance to a less experienced player, positive reinforcement (verbal/written encouragement or compliments).\\
    \textbf{Helping the community:} e.g., encouraging others to report disruptive behavior, submitting validated reports for disruptive behavior, standing up for another player.\\
    \textbf{Celebrating with other players}\\
    \textbf{Congratulating other players}\\
    \textbf{Thanking other players}\\
    \textcolor{ForestGreen}{\underline{Helping with game equipment setup.}}\\
    \textcolor{red}{
    Exclude only requesting help and support without offering any. \\
    Exclude simply pointing out actions that can negatively affect team cohesion.\\
    \underline{Exclude accusing others of not being prosocial or questioning their motives.} \\
    \underline{Exclude threatening to abandon own team, aid other team.} \\
    \underline{Exclude being toxic to own team.}}
    \end{tcolorbox}
    \\
    \bottomrule
  \end{tabularx}
\end{table*}

\begin{table*}[ht]
  \centering
  \caption{\textbf{Evolution of the Definition of Prosocial Behavior Used in Evaluation - Part 2.} Cumulative added exclusions are marked in \textcolor{red}{red} and inclusions in \textcolor{ForestGreen}{green}. Changes made from precious definition iteration are \underline{undelined}.}
  \vspace{0.5em}
  \label{tab:prosocial_definitions_b}
  \renewcommand{\arraystretch}{1.4}
  \begin{tabularx}{\textwidth}{>{\bfseries}l|X}
    \toprule
    Def. ID & Description \\
    \midrule
     D5 & 
    \begin{tcolorbox}[
      colback=white,
      colframe=white,
      boxrule=0.3pt,
      arc=0pt,
      outer arc=0pt,
      left=4pt, right=4pt, top=4pt, bottom=-6pt,
      sharp corners
    ]
    Any action or interaction that benefits other players or the community by fostering a positive gaming environment.\\
    Non-exhaustive list of examples: \\
    \textbf{Cooperating with other players:} e.g., strategizing with team, spawning with team, assisting, pinging for valuable loot or nearby enemies, defending the team.\\
    \textbf{Sharing with other players:} e.g., resources, revives, rewards/gifts.\\
    \textbf{Helping other players:} e.g., providing guidance to a less experienced player, positive reinforcement (verbal/written encouragement or compliments).\\
    \textbf{Helping the community:} e.g., encouraging others to report disruptive behavior, submitting validated reports for disruptive behavior, standing up for another player.\\
    \textbf{Celebrating with other players}\\
    \textbf{Congratulating other players}\\
    \textbf{Thanking other players}\\
    \textcolor{ForestGreen}{Helping with game equipment issues \underline{to facilitate gameplay.}}\\
    \textcolor{red}{
    Exclude only requesting help and support without offering any, \underline{especially repeated nagging for help.} \\
    Exclude simply pointing out actions that can negatively affect team cohesion.\\
    Exclude accusing others of not being prosocial or questioning their motives. \\
    Exclude threatening to abandon own team, aid other team. \\
    Exclude being toxic to own team.}
    \end{tcolorbox}
    \\
    \midrule
     D6 & 
    \begin{tcolorbox}[
      colback=white,
      colframe=white,
      boxrule=0.3pt,
      arc=0pt,
      outer arc=0pt,
      left=4pt, right=4pt, top=4pt, bottom=-6pt,
      sharp corners
    ]
    Any action or interaction that benefits other players or the community by fostering a positive gaming environment.\\
    Non-exhaustive list of examples: \\
    \textbf{Cooperating with other players:} e.g., strategizing with team, spawning with team, assisting, pinging for valuable loot or nearby enemies, defending the team.\\
    \textbf{Sharing with other players:} e.g., resources, revives, rewards/gifts.\\
    \textbf{Helping other players:} e.g., providing guidance to a less experienced player, positive reinforcement (verbal/written encouragement or compliments).\\
    \textbf{Helping the community:} e.g., encouraging others to report disruptive behavior, submitting validated reports for disruptive behavior, standing up for another player.\\
    \textbf{Celebrating with other players}\\
    \textbf{Congratulating other players}\\
    \textbf{Thanking other players}\\
    \textcolor{ForestGreen}{Helping with game equipment issues to facilitate gameplay.}\\
    \textcolor{red}{
    Exclude only requesting help and support without offering any. \\
    \underline{Exclude extended pleading and reiteration of the request bordering on nagging and desperation clearly flooding the chat.} \\
    Exclude simply pointing out actions that can negatively affect team cohesion.\\
    Exclude accusing others of not being prosocial or questioning their motives. \\
    \underline{Exclude excessive complaints about own team, or complaining about lack of help from others.} \\
    Exclude threatening to abandon own team, aid other team. \\
    Exclude being toxic to own team.}
    
    \end{tcolorbox}
    \\

    \bottomrule
  \end{tabularx}
\end{table*}

\section{Calibration Plots for Base Model Selection}
\label{apx:calibration_vsm_bert}

\begin{figure*}[h!]
  \centering
  \includegraphics[width=\textwidth]{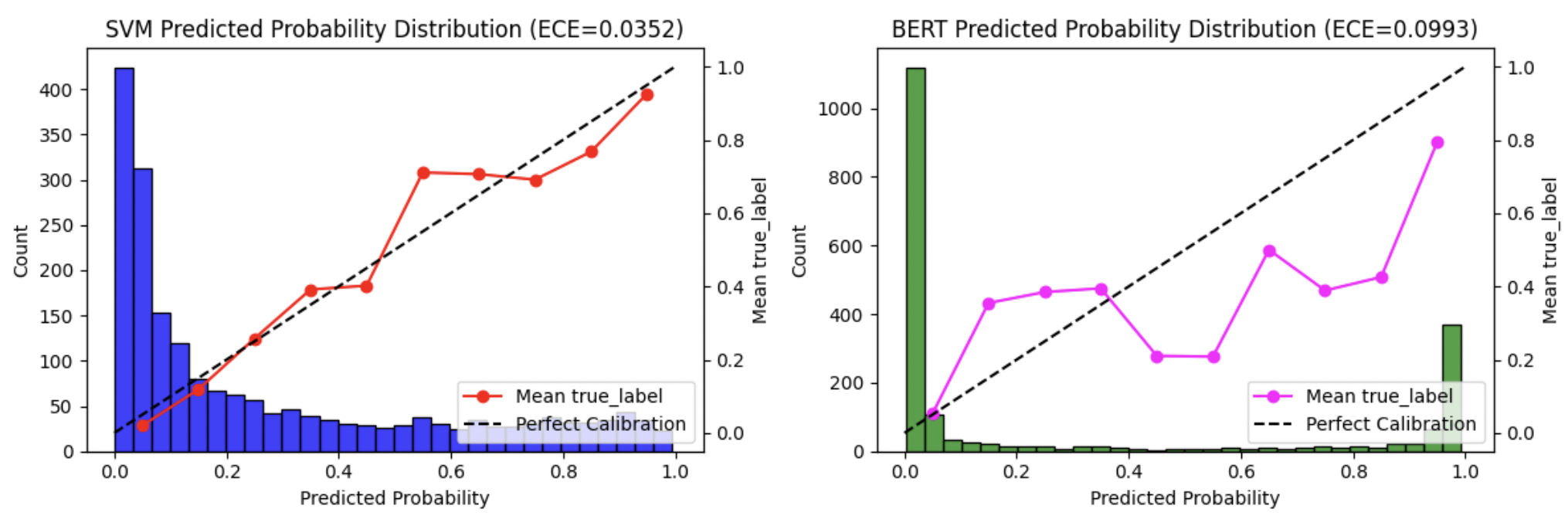}
  \caption{
  \textbf{Calibration of Predicted Probabilities for SVM and BERT Classifiers.}
  The histogram bars (blue for SVM, green for BERT) show the distribution of predicted probabilities across the test set. Overlaid lines show bin-wise mean true labels (red and magenta) compared to the diagonal line of perfect calibration. SVM achieves strong calibration (ECE = 0.0352), while BERT shows significant miscalibration (ECE = 0.0993), especially in mid-probability regions. Well-calibrated scores are essential for downstream routing in hybrid inference pipelines.}
  \label{fig:apx_calib_plots}
\end{figure*}

Figure~\ref{fig:apx_calib_plots} compares calibration for SVM and BERT models via predicted probability distributions and reliability curves. SVM exhibits strong calibration (ECE = 0.0352), closely following the ideal diagonal. BERT shows substantial miscalibration (ECE = 0.0993), particularly in mid-confidence regions.

Both models produce many low-confidence predictions, but BERT tends to overconfident extremes, misaligned with true label frequencies. These findings support selecting SVM as a calibration-stable base classifier for cost-efficient hybrid pipelines.

\end{document}